\documentclass[draftcls,onecolumn]{IEEEtran}

\usepackage{amsmath}
\usepackage{bm}

\usepackage[ruled]{algorithm2e}

\usepackage{cite}

\usepackage{graphicx}
\usepackage{subfigure}
\usepackage[justification=centering]{caption} 

\usepackage{multirow}
\usepackage{booktabs}

\usepackage{tikz}
\usepackage{comment}
\usepackage{amsmath,amssymb} 
\usepackage{color}

\ifCLASSOPTIONcompsoc
  \usepackage[nocompress]{cite}
\else
  \usepackage{cite}
\fi

\begin{document}

\title{Gated Domain-Invariant Feature Disentanglement for Domain Generalizable Object Detection}

\author{\IEEEauthorblockN{Haozhuo~Zhang}
\IEEEauthorblockA{haozhuozhang@zju.edu.cn}\\
\and
\IEEEauthorblockN{Huimin~Yu}
\IEEEauthorblockA{yhm2005@zju.edu.cn}\\
\and
\IEEEauthorblockN{Yuming~Yan}
\IEEEauthorblockA{21931092@zju.edu.cn}\\
\and
\IEEEauthorblockN{Runfa~Wang}
\IEEEauthorblockA{21960189@zju.edu.cn}
}

\IEEEtitleabstractindextext{%
\begin{abstract}
For Domain Generalizable Object Detection (DGOD), Disentangled Representation Learning (DRL) helps a lot by explicitly disentangling Domain-Invariant Representations (DIR) from Domain-Specific Representations (DSR). 
Considering the domain category is an attribute of input data, it should be feasible for networks to fit a specific mapping which projects DSR into feature channels exclusive to domain-specific information, and thus much cleaner disentanglement of DIR from DSR can be achieved simply on channel dimension.
Inspired by this idea, we propose a novel DRL method for DGOD, which is termed Gated Domain-Invariant Feature Disentanglement (GDIFD). In GDIFD, a Channel Gate Module (CGM) learns to output channel gate signals close to either $0$ or $1$, which can mask out the channels exclusive to domain-specific information helpful for domain recognition. 
With the proposed GDIFD, the backbone in our framework can fit the desired mapping easily, which enables the channel-wise disentanglement.
In experiments, we demonstrate that our approach is highly effective and achieves state-of-the-art DGOD performance.
\end{abstract}

}

\maketitle

\IEEEdisplaynontitleabstractindextext

\IEEEpeerreviewmaketitle

\ifCLASSOPTIONcompsoc
\IEEEraisesectionheading{\section{Introduction}\label{intro}}
\else
\section{Introduction}
\label{intro}
\fi

As a method for locating objects and identifying target categories, object detection plays a fundamental and vital role in computer vision. The past decade has witnessed the booming development of supervised object detection, many great works have demonstrated excellent performance when the training and testing data are collected from the same domain~\cite{rcnn,fastrcnn,fasterrcnn,yolo,ssd,lin2017focal,maskrcnn,hu2018relation,tian2019fcos}. However, labels of the target domain are probably difficult to obtain, and in some cases target domain data is even unavailable for training. Due to the domain gap between target domain and source domain, the absence of target labels or even target data for training will lead to performance degradation of conventional detectors on target domain.
To address this problem, an emerging {\it Domain Generalizable Object Detection} (DGOD) method~\cite{didn} and many {\it Unsupervised Domain Adaptive Object Detection} (UDAOD) methods~\cite{chen2018domainWild,he2019multiGRL,zhuang2020ifan,cfa,wang2021domain,rezaeianaran2021seeking,vdd,liu2022decompose} have been proposed to enhance the generalizability of object detectors. The DGOD method only requires labeled data from several source domains to align the data distributions and achieves better detector generalizability on the unseen target domain. UDAOD methods, as less powerful ways, utilize both labeled source data and unlabeled target data to reduce the domain gap. In this paper, we focus on the DGOD task that is more challenging.
\begin{figure}
	\centering
	\includegraphics[height=5.1cm]{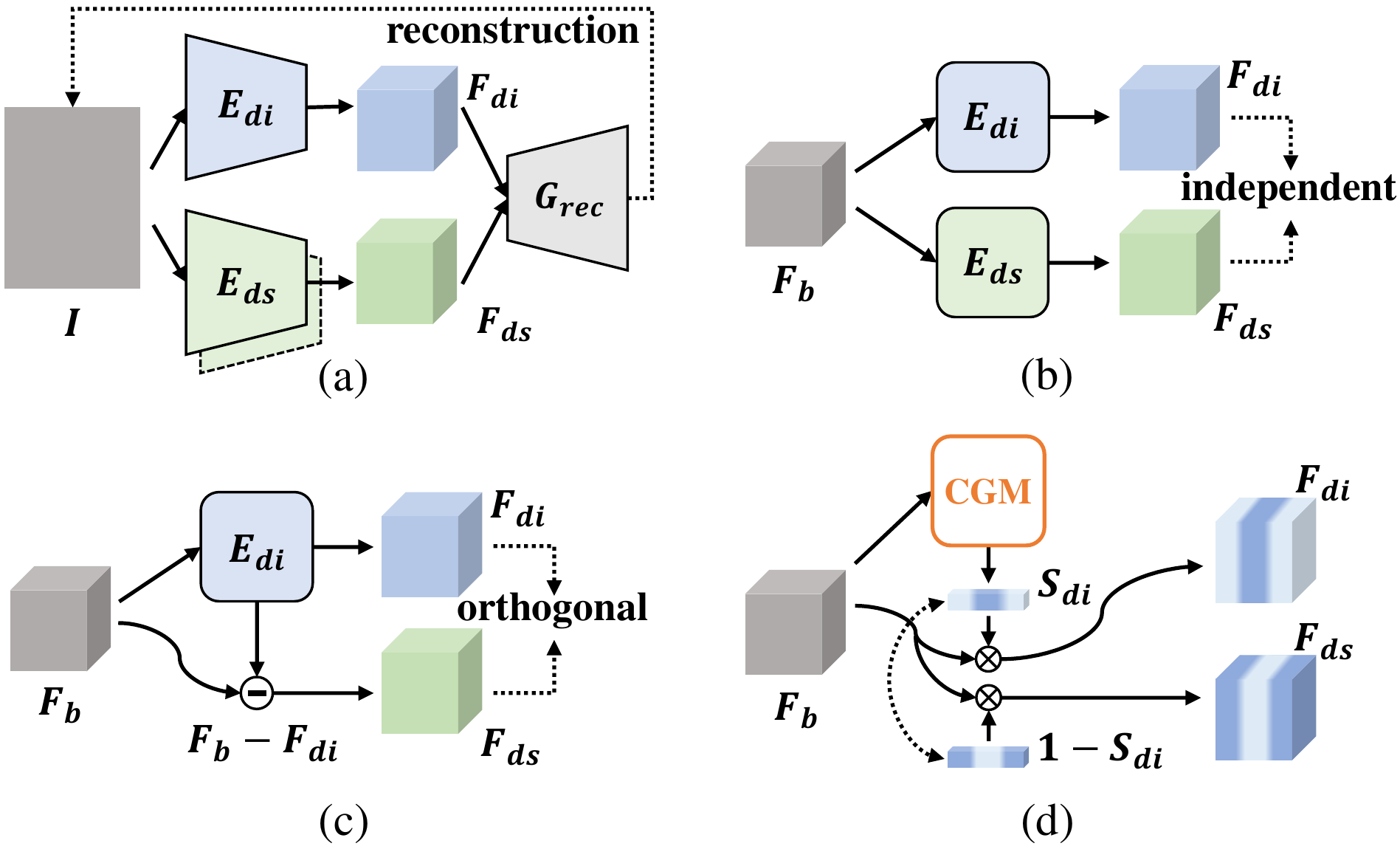}
	\caption{(a) Using a shared domain-invariant backbone $E_{di}$ across domains and one domain-specific backbone $E_{ds}$ for each source domain to extract DIR $F_{di}$ and DSR $F_{ds}$ respectively. A decoder $G_{rec}$ is used to reconstruct the input image $I$. (b) Both $E_{di}$ and $E_{ds}$ are shared shallow encoders following the backbone, specific losses can be utilized to make $F_{di}$ and $F_{ds}$ independent. (c) Only a shared shallow encoder $E_{di}$ is used to extract $F_{di}$ from the backbone feature $F_{b}$, $F_{ds}$ is simply the difference between $F_{b}$ and $F_{di}$. $F_{di}$ and $F_{ds}$ are constrained to be orthogonal. (d) Proposed gated DRL module based on the channel-wise gate signal $S_{di}$ output by Channel Gate Module (CGM)}
	\label{fig:exist_disens}
\end{figure}

The first and currently the only DGOD method DIDN~\cite{didn} employs {\it Disentangled Representation Learning} (DRL) to address DGOD, which attempts to disentangle {\it Domain-Invariant Representations} (DIR) from {\it Domain-Specific Representations} (DSR) explicitly and the feature alignment across domains is performed on the disentangled DIR without the interference from DSR. DGOD can be tackled by detecting objects based on the DIR which has little domain-specific information detrimental to detector generalizability. Specifically, DIDN uses multiple non-shared domain-specific backbones $E_{ds}$ and one shared domain-invariant backbone $E_{di}$ as shown in Fig.~\ref{fig:exist_disens}(a). 
Besides this DRL structure with a huge number of parameters for DGOD, there are two types of lightweight DRL modules in existing UDAOD researches: one is adopted by the powerful UDAOD method DDF~\cite{liu2022decompose} (Fig.~\ref{fig:exist_disens}(b)), the other is the Vector-Decomposed Disentanglement (VDD) proposed in~\cite{vdd} (Fig.~\ref{fig:exist_disens}(c)). These two DRL methods use very shallow encoders $E_{di}$ to extract DIR $F_{di}$ from the backbone feature $F_{b}$, and VDD even directly takes the difference between $F_b$ and $F_{di}$ as $F_{ds}$, which ensures no loss of information in $F_b$ without the reconstruction operation used in DIDN. 
However, a helpful a priori knowledge is not leveraged in designing DRL modules for DGOD and UDAOD. Considering that \textit{the domain category is an attribute of input data}, it is feasible to make specific feature channels responsible for encoding only the domain-specific information related to the domain category, and cleaner disentanglement of DIR from DSR can be achieved by simply filtering out the channels exclusive to domain-specific information. Although domain-specific information can be exclusively projected into specific feature channels by stacked convolutional layers used in existing DRL methods, these vanilla networks do not guarantee to fit the special mapping we described.

Instead of expecting stacked convolutional layers to fit the desired mapping, we propose a novel DRL method Gated Domain-Invariant Feature Disentanglement (GDIFD) to explicitly let the backbone fit the mapping and achieve domain-invariant feature disentanglement simply along the channel dimension. 
Specifically, a Channel Gate Module (CGM) is designed to output channel-wise domain-invariant gate signal $S_{di}$ which is expected to mask out the dedicated channels for domain-specific information and maintains the other domain-invariant ones for object detection. 
As shown in Fig.~\ref{fig:exist_disens}(d), the disentanglement of DIR from DSR will be achieved easily by performing element-wise product of $S_{di}$ and the backbone feature $F_b$. 
To explicitly filter out the domain-specific information on channel dimension, DSR is used to perform domain recognition following our idea. Additionally, a gate loss is proposed to make CGM output $S_{di}$ with values very close to either $0$ or $1$ to enable much cleaner channel-wise disentanglement, which is also the key difference between our channel-wise gate signal and the conventional channel attention~\cite{senet} with unconstrained values ranged from $0$ to $1$. Domain-specific channels are facilitated to be exclusive to DSR by the gate loss, which enables the desired cleaner channel-wise disentanglement. Furthermore, the commonly used adversarial alignment of DIR is adopted to promote the independence between DIR and DSR.
With the help of this novel DRL method, the backbone in our DGOD framework learns to make some channels exclusively encode all the domain information and thus enable simple channel-wise disentanglement.

Our main contributions are summarized as follows:
(1) We propose a novel Gated Domain-Invariant Feature Disentanglement (GDIFD) for DGOD, which enables much cleaner disentanglement on channel dimension. 
(2) A Channel Gate Module (CGM) is designed for better channel-wise gate signal prediction. In addition, a special initialization of CGM is proposed to further improve performance of our DGOD framework.
(3) Compared to the baseline, significant DGOD performance gain of our DRL method is demonstrated in extensive experiments. Additionally, the superiority of GDIFD over other DRL methods and the effectiveness of our module design are validated.

\section{Related Work}

\subsection{DGOD and UDAOD based on Feature Alignment}

Domain gap can be reduced by minimizing the discrepancy of the feature spaces of multiple domains. Ganin \textit{et al.}~\cite{ganin2015unsupervised} first propose the widely used Gradient Reversal Layer (GRL) to enable feature alignment via adversarial learning. The model equipped with GRL can learn to extract domain-invariant features by deceiving an additional domain classifier in training. 
Lin \textit{et al.}~\cite{didn} propose the first and currently the only DGOD method DIDN, they resort to feature disentanglement on both image and instance level to obtain the final domain-invariant instance-level feature for detection. The disentanglement is promoted by adversarially aligning the domain-invariant features. Due to the separation of domain-specific features, the alignment becomes easier and thus DIDN achieves good performance.
As for the UDAOD task, Chen \textit{et al.}~\cite{chen2018domainWild} first propose an end-to-end UDAOD method by applying adversarial feature alignment on image and instance level. Saito \textit{et al.}~\cite{strongweak} introduce adaptive weights for domain classification losses to pay more attention on the locations which are easier for domain confusion. By this way, the difficulty of adversarial feature alignment is reduced. Zheng \textit{et al.}~\cite{zheng2020cross} use attention maps to make adversarial alignment focus on the foreground areas. Additionally, they build a prototype for each category in different domains and achieve alignment by explicitly drawing prototypes of the same category closer to each other. Considering the noise from background, Hsu \textit{et al.}~\cite{cfa} conduct center-aware feature alignment by using the output centerness and classification maps of FCOS~\cite{tian2019fcos} detector as weights for the features to align. Guan \textit{et al.}~\cite{uadan} further introduce an uncertainty metric to focus on the pool-aligned samples to achieve more refined and robust feature alignment.

\subsection{DRL in Domain Generalization and Domain Adaptation}
Besides the aforementioned DGOD method DIDN, DRL is widely used for DG and DA tasks because of the ability to ease feature alignment by explicitly disentangling DIR from DSR. 

Peng \textit{et al.}~\cite{peng2019domain} extend the adversarial training to the classification task and propose a novel three-branch DRL structure which can output class-irrelevant features, domain-specific features and desired domain-invariant features which are also task-relevant. In this way, more noise is separated. Nonetheless, this method requires additional reconstruction like DIDN does. Chen \textit{et al.}~\cite{chen2021style} use a simpler DRL structure as depicted in Fig.~\ref{fig:exist_disens}(b) for DG classification. A novel domain-specific style memory is built to constrain the style in a single domain to be consistent, which facilitate the DRL. Liu \textit{et al.}~\cite{liu2022decompose} use the same DRL method as the former algorithm and achieves remarkable UDAOD results by increasing the distances between DIR and DSR on both image and instance level. In addition, the performance can be further enhanced by reducing the similarity between instance features from different domains. Wu \textit{et al.}~\cite{vdd} propose a novel vector-decomposed DRL structure which only has one extractor for DIR and uses simple orthogonal constraint instead of reconstruction (Fig.~\ref{fig:exist_disens}(c)). 

Different from the existing DRL ways, we note that it is possible to disentangle DIR from DSR on channel dimension. We propose a gated DRL method to explicitly make the backbone project information private for domains into some dedicated channels. By this way, channel-wise domain-invariant feature disentanglement can be achieved simply by only maintaining the complementary domain-invariant channels of a backbone feature.


\section{Proposed Method}

\subsection{Algorithm Overview}
There are sufficient available labeled data from multiple source domains during training for DGOD, and we aim to use these data to train an object detector that generalizes well on the unseen target domain. To achieve this goal, we employ Disentangled Representation Learning (DRL) to explicitly disentangle the domain-invariant feature $F_{di}$ from the domain-specific feature $F_{ds}$ that has negative effects on model generalization. 
As shown in Fig.~\ref{fig:model}, we design a novel Gated Domain-Invariant Feature Disentanglement (GDIFD) module which is shared for the five-scale features extracted by the backbone $E_b$ with Feature Pyramid Network (FPN)~\cite{fpn}. For a deep feature $F_b$, the proposed Channel Gate Module (CGM) outputs a channel-wise gate signal $S_{di}$ which can mask out the channels exclusive to domain-specific information. 
The DIR $F_{di}$ we desire can be obtained by doing element-wise product for $F_b$ and $S_{di}$, and the complementary DSR $F_{ds}$ is the element-wise product of $F_b$ and $\textbf{1}-S_{di}$. The explicit extraction of DSR on channel dimension is achieved by domain recognition via the domain classification loss $\mathcal{L}_{D-cls}$. 
In addition, a gate loss $\mathcal{L}_{gate}$ calculated using $S_{di}$ and $\textbf{1}-S_{di}$ facilitates the CGM to output gate signal with values close to $0$ or $1$ and enables much cleaner disentanglement on channel dimension.
To further promote the independence between DIR and DSR, the adversarial domain classification loss $\mathcal{L}_{D-adv}$ is imposed on DIR, which can largely remove the domain-specific information in DIR.
Finally, the detection results are output by the FCOS head $H$ fed with the domain-invariant feature $F_{di}$.

\subsection{Gated Domain-Invariant Feature Disentanglement}
\label{gdifd}
\begin{figure}
	\centering
	\includegraphics[height=3.9cm]{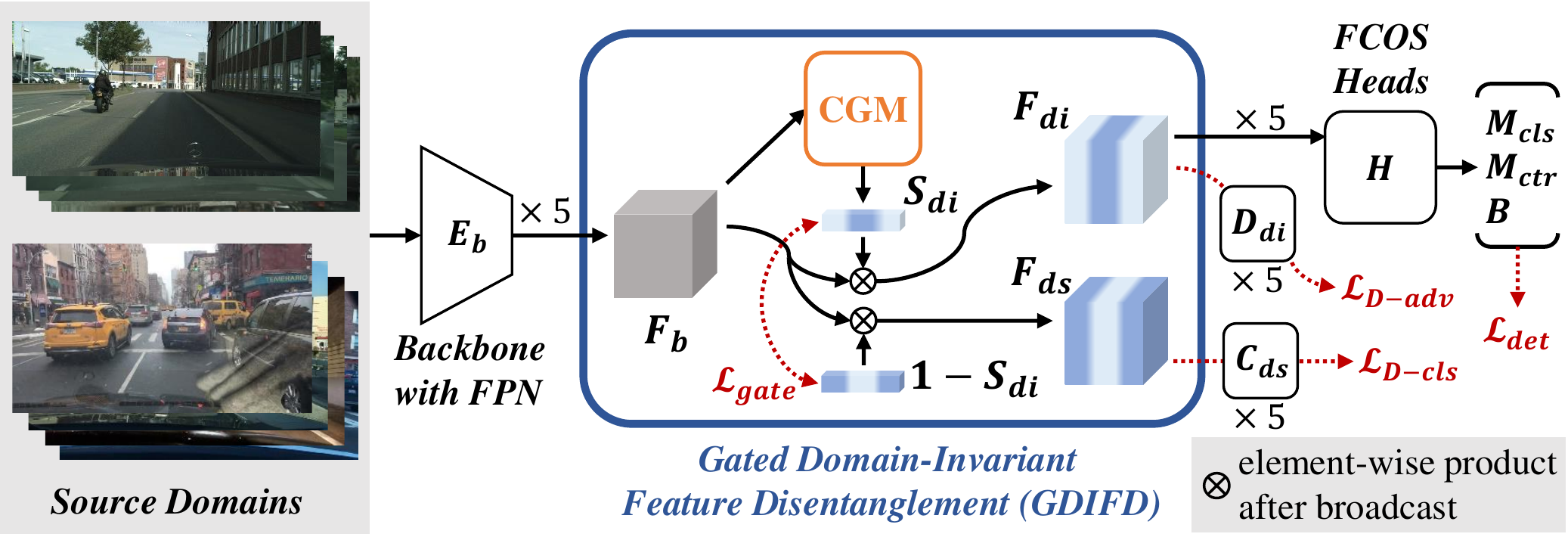}
	\caption{A baseline model equipped with the proposed Gated Domain-Invariant Feature Disentanglement (GDIFD) for DGOD. Images from source domains are fed to a shared backbone network $E_b$ which outputs deep features of 5 scales. The proposed GDIFD module and FCOS heads $H$ are shared across the 5 scales. Adversarial feature alignment using loss $\mathcal{L}_{D-adv}$ and domain classification using loss $\mathcal{L}_{D-cls}$ are imposed on DIR $F_{di}$ and DSR $F_{ds}$, respectively. $D_{di}$ and $C_{ds}$ are the discriminator with GRL and the domain classifier, respectively. The proposed gate losses $\mathcal{L}_{gate}$ are computed using channel gate signal $S_{di}$. Detection losses $\mathcal{L}_{det}$ are computed using outputs of $H$}
	\label{fig:model}
\end{figure}

To enable the novel channel-wise disentanglement, we first design the CGM (described in Section~\ref{cgm}) for GDIFD to output the channel gate signal $S_{di} \in \mathbb{R}^{C}$ whose values ranged from $0$ to $1$. With the help of this channel gate signal, a backbone feature $F_b \in \mathbb{R}^{C\times H\times W}$  of $C$ channels and $H \times W$ spatial resolution can be separated into two complementary parts denoted as $F_{di} \in \mathbb{R}^{C\times H\times W}$ and $F_{ds} \in \mathbb{R}^{C\times H\times W}$ by performing element-wise product of $F_b$ with $S_{di}$ and $\textbf{1}-S_{di}$ respectively as shown in Fig.~\ref{fig:model}.

As depicted in Section~\ref{intro}, we propose to achieve the disentanglement of DIR from DSR on channel dimension by utilizing the a priori knowledge that domain category is an attribute of the input data. Specifically, $F_{ds}$ desired to be DSR is fed to a five-layer fully convolutional domain classifier $C_{ds}$ to conduct domain classification for each spatial location on $F_{ds}$. The domain category label is set as $T_s \in \mathbb{R}^{M}$ for a source domain $S_s \in \{S_1, S_2, \dots, S_M\}$, where $M$ is the number of the source domains can be seen. The value $T_s^{(s)}$ is $1$ and $T_s^{(d)}$ is $0$ for $d\neq s$. For an input image $I_s$ from the source domain $S_s$, the domain classification loss is defined as:
\begin{align}
	\mathcal{L}_{D-cls}(I_s) & = -\dfrac{1}{H\times W} \sum_{u=1,v=1}^{H,W} \sum_{d=1}^{M} (T_s^{(d)}\log(C_{ds}(F_{ds})^{(u,v,d)}))\ .
\end{align}

To achieve much cleaner disentanglement on channel dimension, we further try to let the backbone fit a special mapping that projects the encoded domain-specific information into channels \textit{exclusive} to DSR. A gate loss which constrains the values of $S_{di}$ very close to either $0$ or $1$ is designed to enable the cleaner channel-wise disentanglement. The gate loss is calculated as:
\begin{align} \label{eq:loss_gate}
	\mathcal{L}_{gate} & = {\rm max} (S_{di}^{T}(\textbf{1}-S_{di}) - m, 0)\ .
\end{align}

$m$ is a margin we add to ease the training, since the gate signal output by sigmoid activation are in the range of $(0, 1)$ and the inner product of $S_{di}$ and $\textbf{1}-S_{di}$ can never be $0$. We simply set $m$ as $0.01$. With the help of $\mathcal{L}_{gate}$, the orthogonality desired for DRL between $F_{di}$ and $F_{ds}$ can be met easily in GDIFD, since the inner product of $F_{di}^{(u,v)}$ and $F_{ds}^{(u,v)}$ for a spatial location $(u,v)$ will be a positive scalar near $0$ with all the values of $S_{di}$ very close to either $0$ or $1$. 

At this point, the unwanted DSR is extracted explicitly and the DIR mainly encodes the information for the object detection task.
However, it should be noted that the orthogonality does not guarantee $F_{di}$ and $F_{ds}$ are independent. To promote the independence between DIR and DSR, the commonly used \textit{adversarial feature alignment method}~\cite{chen2018domainWild} is imposed on $F_{di}$ to ensure the domain-specific information is largely removed from DIR.
Specifically, $F_{di}$ is fed to a discriminator $D_{di}$ with the same structure of $C_{ds}$ to perform domain classification. For an input image $I_s$ from the source domain $S_s$, the adversarial loss function for an extracted $F_{di}$ is defined as:
\begin{align} \label{eq:loss_adv}
	\mathcal{L}_{D-adv}(I_s) & = -\dfrac{1}{H\times W} \sum_{u=1,v=1}^{H,W} \sum_{d=1}^{M} (T_s^{(d)}\log(D_{di}(F_{di})^{(u,v,d)}))\ .
\end{align}

A Gradient Reversal Layer (GRL)~\cite{ganin2015unsupervised} which reverses the sign of the backpropagating gradient is inserted between the extractor of $F_{di}$ and the discriminator $D_{di}$ to make the domain-invariant part of our model learn to adversarially deceive $D_{di}$ by outputting the desired $F_{di}$ with little domain-specific information for domain recognition. 

\subsection{Channel Gate Module}
\label{cgm}
As described in Section~\ref{gdifd}, the CGM is designed to output channel-wise signal $S_{di} \in \mathbb{R}^{C}$.
The things that come to mind firstly when we think of channel-wise predictions are likely to be the widely used channel attention which can be output by SE block~\cite{senet} and CBAM~\cite{cbam}. With these channel attention modules, $S_{di}$ of the same shape as the channel attention can certainly be obtained. However, these modules are subject to efficiency constraints. Specifically, the global pooling operation before encoding is so rough that the details of domain-level information may be lost, which will increase the difficulty and instability of the subsequent gate signal prediction. Instead, we insert the global average pooling layer \textit{after} per-pixel gate signal prediction in the proposed CGM to let every feature location ``vote" for the final global channel signal $S_{di}$. 

\begin{figure}
	\centering
	\includegraphics[height=3.3cm]{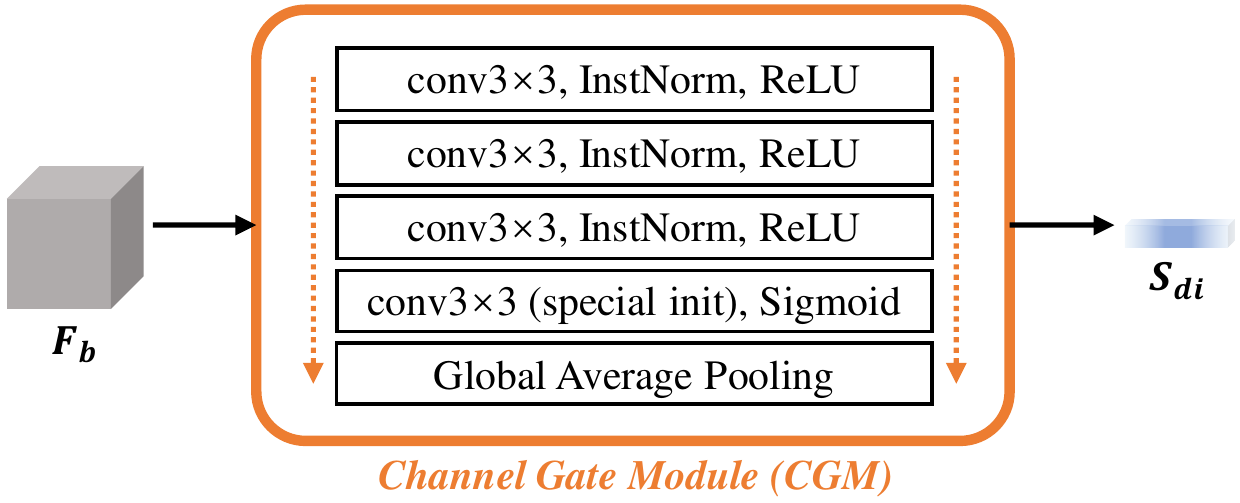}
	\caption{Network design of the proposed Channel Gate Module (CGM). The bias of the last convolutional layer is specially initialized in training}
	\label{fig:cgm}
\end{figure}

Still different from usual methods, we use four $3\times 3$ convolutional layers of stronger representational capacity followed by Instance Normalization~\cite{instnorm} and ReLU activation~\cite{relu} to predict the channel signal for each feature location (Fig.~\ref{fig:cgm}). Note that the last convolutional layer is followed by sigmoid for outputting signal values ranged from $0$ to $1$. 

In addition, the task-relevant DIR should be much more informative than the DSR encoding only domain-specific information for domain recognition and thus the optimal backbone feature $F_b$ may have fewer channels dedicated to DSR encoding. Based on this thought, we try to make all the values of $S_{di}$ very close to $1$ at the start of training, which is probably close to the optimal case and can help our DGOD framework achieve better performance.
Specifically, the special initialization is achieved by setting the bias of the last convolutional layer as large values to get $s_{init}$ close to $1$, since the weights of convolutional layers are usually initialized to values close to $0$ and $s_{init}$ is close to $Sigmoid(b_{init})$. $b_{init}$ denotes the initial value filling the bias of the last convolutional layer and $s_{init}$ denotes the expected initial values filling $S_{di}$. Then we can calculate $b_{init}$ as:
\begin{align}
	b_{init} & = -\log(\dfrac{1-s_{init}}{s_{init}})\ .
	\label{eq:b_init}
\end{align}

\subsection{Overall Objective for Proposed DGOD Framework} \label{losses}
Following the UDAOD method CFA~\cite{cfa}, our DGOD framework is based on the FCOS detector~\cite{tian2019fcos}. As shown in Fig.~\ref{fig:model}, the backbone with FPN outputs features of five scales, which can be denoted as $F_b^l \in \mathbb{R}^{C\times H\times W}$ with $l\in \{3,4,5,6,7\}$ and $C=256$. Our disentangled module and the FCOS head are shared across the five scale levels, but the domain discriminators and domain classifiers depicted in Section~\ref{gdifd} for feature alignment are not shared.

As for the detection task, for a backbone feature $F_b^l$ of an input image $I_s$, the fully convolutional FCOS head will output an object classification score map $M_{cls}$, a centerness map $M_{ctr}$ predicting the distances of each pixel to the four sides of the corresponding bounding box and a regression map $B$ containing the encoded predicted bounding box information. Given the ground-truth bounding boxes $B_{gt}$, loss functions $\mathcal{L}_{cls}, \mathcal{L}_{ctr}$ and $\mathcal{L}_{reg}$ will be calculated for the three FCOS head outputs respectively. $\mathcal{L}_{cls}$ is the focal loss~\cite{lin2017focal}, $\mathcal{L}_{ctr}$ is the binary cross-entropy loss and $\mathcal{L}_{reg}$ is the IoU loss~\cite{unitbox}. The loss function $\mathcal{L}$ for the detection task of scale level $l$ can be written as:
\begin{align}
	\mathcal{L}_{det}^l(I_s, B_{gt}) & = \mathcal{L}_{cls}^l + \mathcal{L}_{ctr}^l + \mathcal{L}_{reg}^l\ .
\end{align}

Denoting the previously depicted losses of scale level $l$ for the GDIFD module as $\mathcal{L}_{D-adv}^l,\mathcal{L}_{D-cls}^l$ and $\mathcal{L}_{gate}^l$, the overall loss function can be expressed as:
\begin{align}
	\mathcal{L}(I_s, B_{gt}) & = \sum_{l=3}^{7}(\mathcal{L}_{det}^l + \lambda_{D-adv}\mathcal{L}_{D-adv}^l + \lambda_{D-cls}\mathcal{L}_{D-cls}^l + \lambda_{gate}\mathcal{L}_{gate}^l)\ .
\end{align}

\section{Experiment}
\subsection{Datasets and DGOD settings}
The dataset settings described in the first and currently the only DGOD method DIDN~\cite{didn} are followed and there are five datasets used to simulate DGOD in the real world. 

\subsubsection{Datasets.}
The following five large-scale object detection datasets are used: \textit{BDD100k}~\cite{bdd100k}, \textit{Cityscapes}~\cite{cityscapes}, \textit{Foggy Cityscapes}~\cite{cityscapes}, \textit{KITTI}~\cite{kitti} and \textit{SIM 10k}~\cite{sim10k}. 

BDD100k is a large-scale dataset which is collected by a driving platform on streets. This dataset is very diversified but only the validation set including $10000$ images is used for DGOD simulation. Cityscapes is a dataset of urban scenes for driving scenarios and Foggy Cityscapes is the synthetic foggy version of Cityscapes. These two dataset both have $2975$ images in the training set and $500$ images in the validation set. KITTI is a database for autonomous driving, which includes $7481$ images. SIM 10k is a virtual dataset containing $10000$ images rendered by the GTA gaming engine.
For simpler representations of different DGOD settings, we will denote the five datasets as B, C, F, K, and S, respectively.

\subsubsection{DGOD settings.}
There are eight categories with instance labels shared across BDD100k, Cityscapes and Foggy Cityscapes. But KITTI and SIM 10k only have the car category. Following DIDN~\cite{didn}, we first conduct DGOD experiments based on the shared eight categories using BDD100k, Cityscapes and Foggy Cityscapes. Specially, the ``train" category is not evaluated for BDD100k because there are few train objects in this dataset. The second setting is to utilize all the datasets and only considering the shared ``car" category. Following DIDN, BDD100k and KITTI is used as the validation dataset to evaluate a DGOD method respectively in this setting. The ablation study is conducted in the multi-category DGOD settings following DIDN.

\subsection{Implementation Details}
\label{exp_detail}
Our method is implemented with PyTorch~\cite{pytorch}. We use FCOS~\cite{tian2019fcos} without generalization operations as our baseline, which is denoted as ``Base" for simplicity.
We empirically set $\lambda_{D-adv}$, $\lambda_{D-cls}$, $\lambda_{gate}$ and $s_{init}$ as $0.1$, $0.01$, $0.1$ and $0.9999$ respectively in experiments, unless specified. The weight for the reversed gradient of GRL is set as $0.01$. ResNet-50~\cite{resnet} and ResNet-101~\cite{resnet} pre-trained on ImageNet~\cite{imagenet} with FPN~\cite{fpn} is used as the backbone of the FCOS detector. Following CFA, the input images are resized with their shorter side as $800$ and longer side less or equal to $1333$. We use SGD optimizer with momentum of $0.9$ and weight decay of $1\times 10^{-4}$.
\setlength{\tabcolsep}{4pt}
\begin{table}
	\begin{center}
		\caption{Results (\%) of three DGOD settings using B, C and F}
		\label{table:setting3}
		\begin{tabular}{clccccccccc}
			\hline \noalign{\smallskip} 
			& Method   &person& car & train & rider & truck & motor & bike & bus & mAP \\
			\hline \noalign{\smallskip}
			\multirow{6}*{B\&C} & DIDN-50 & 31.8 & 49.3 & \textbf{26.5} & 38.4 & 27.7 & 24.8 & 33.1 & 35.7 & 33.4 \\
			\multirow{6}*{toF}  & Base-50 & 39.8 &  58.2 & 4.7 & 39.5 & 26.7 & \textbf{28.5} & 36.7 & 37.2 &  33.9 \\
			& Ours-50  & 41.1 & 59.5 & 19.7 & 40.2 & 28.3 & 28.2 & 35.9 & \textbf{41.3} & 36.8 \\
			& Base-101 & 40.1 & 59.5 & 7.2 & 39.5 & 29.8 & 27.9 & 37.3 & 38.2 & 35.0\\
			& Ours-101 & \textbf{41.5} & \textbf{59.7} & 25.4 & \textbf{42.2} & \textbf{31.8} & 27.8 & \textbf{38.6} & 39.4 & \textbf{38.3} \\ 
			\cmidrule{2-11}
			& Oracle-101   & 49.0 & 68.0 & 42.9 & 46.2 & 35.0 & 35.4 & 41.7 & 56.0 & 46.8 \\
			
			\hline \noalign{\smallskip} 
			\multirow{6}*{B\&F} & DIDN-50     & 43.6 & 63.2 & 51.1 & 46.2 & 41.9 & 36.0 & \textbf{41.3} & \textbf{60.9} & 47.9 \\
			\multirow{6}*{toC}  & Base-50 & 48.7 & 70.6 & 17.2 & 43.3 & 36.2 & 36.0 & 38.1 & 49.5 & 42.5  \\
			& Ours-50  & 50.4 & \textbf{71.2} & 46.8 & 45.5 & 40.3 & \textbf{38.8} & 39.8 & 56.8 & 48.7 \\
			& Base-101 & 49.2 & 70.3 & 24.1 & 43.4 & 43.5 & 34.2 & 38.2 & 51.8 & 44.3  \\
			& Ours-101 & \textbf{50.9} & 70.7 & \textbf{52.1} & \textbf{47.6} & \textbf{46.0} & 37.2 & 39.5 & 60.4 & \textbf{50.6} \\ 
			\cmidrule{2-11}
			& Oracle-101   & 52.1 & 70.6 & 53.4 & 50.6 & 40.5 & 39.8 & 42.6 & 57.6 & 50.9 \\
			
			\hline \noalign{\smallskip} 						  
			\multirow{6}*{C\&F} & DIDN-50 & 34.5 & 44.2 & - & \textbf{30.4} & \textbf{21.2} & 9.2 & \textbf{22.8} & 19.0 & 22.7 \\
			\multirow{6}*{toB}  & Base-50 & 39.0 & 49.3 & - & 21.8 & 15.1 & 9.4 & 21.6 & 14.0 & 21.3 \\
			& Ours-50  & 40.1 & \textbf{51.1} & - & 25.7 & 16.9 & 7.5 & 22.2 & 12.8 & 22.0 \\
			& Base-101 & 36.6 & \textbf{51.1} & - & 23.7 & 16.6 & \textbf{10.7} & 20.2 & 13.2 & 21.6 \\
			& Ours-101     & \textbf{40.3} & \textbf{51.1} & - & 23.6 & 19.8 & 6.5 & 22.1 & \textbf{19.4} & \textbf{22.9} \\ 
			\cmidrule{2-11}
			& Oracle-101   & 55.0 & 73.9 & - & 29.1 & 51.4 & 23.0 & 35.0 & 45.3 & 39.1  \\
			\hline
		\end{tabular}
	\end{center}
\end{table}
\setlength{\tabcolsep}{1.4pt}

There are two training schedules in our experiments. We note that the model performance can converge fast in the DGOD settings ``C\&F to B", ``B\&C\&F\&S to K" and ``C\&F\&K\&S to B". Therefore, we first train with learning rate $0.003$ for $4K$ iterations and then train with learning rate $0.0003$ for another $4K$ iterations to achieve fast training. For all the experiments including the ``Oracle" ones (train and test on the same target domain) in other DGOD settings, the model is trained with learning rate $0.005$ for $14K$ iterations and then trained with learning rate $0.0005$ for another $6K$ iterations. In addition, there is a 500-iteration warm-up stage with very small learning rate in the beginning of training for all experiments. One GPU is used for one session of training and test. In one iteration, we iterate over the source domains to train the model. For one domain in an iteration, a batch composed of four images from this domain is fed to the model for one step of optimization. Following the first DGOD method DIDN, mean average precision (mAP) with a threshold of 0.5 is employed to evaluate the results of all the categories.

\subsection{Domain Generalization Results}
\subsubsection{Domain Generalization Results.}  \label{main_results}
We report the quantitative results of the state-of-the-art algorithm DIDN, the non-disentangled baseline and our method in Table~\ref{table:setting3} and Table~\ref{table:setting2}. The Oracle results of the vanilla FCOS detector with ResNet-101 on the  multi-category DGOD settings are also reported. We have the following observations based on these results.

(1) Compared with the Oracle results in Table~\ref{table:setting3}, the DGOD performance drops considerably on all the settings except for the one with the simpler validation set from Cityscapes. The negative effect of the domain shift problem can be demonstrated by comparing the DGOD results with the Oracles. 
\setlength{\tabcolsep}{4pt}
\begin{table}
	\begin{center}
		\caption{Results (\%) of two Domain Generalization settings using B, C, F, K and S}
		\label{table:setting2}
		\begin{tabular}{lccccc}
			\hline \noalign{\smallskip} 
			Method 	 & DIDN-50 & Base-50 & Ours-50 & Base-101 & Ours-101 \\
			\hline \noalign{\smallskip}
			B\&C\&F\&S to K    & 76.8 & 77.6 & 79.4 & 78.2 & \textbf{80.1} \\
			C\&F\&K\&S to B    & 52.3 & 54.1 & 55.3 & 55.9 & \textbf{57.0} \\
			\hline
		\end{tabular}
	\end{center}
\end{table}
\setlength{\tabcolsep}{1.4pt}

(2) The state-of-the-art DGOD algorithm DIDN is even inferior to the FCOS baseline by $0.5\%$, $0.8\%$, $1.8\%$ in the settings of B\&C to F, B\&C\&F\&S to K and C\&F\&K\&S to B. However, it should be noted that the remarkable performance of our DGOD framework is largely attributed to the proposed gated DRL method rather than the simple baseline. Our method achieves $2.9\%$, $6.2\%$, $0.7\%$, $1.8\%$ and $1.2\%$ improvement against the baseline with ResNet-50 in Foggy Cityscapes, Cityscapes, multi-category BDD100k, KITTI and single-category BDD100k, respectively. Compared to the baseline equipped with ResNet-101 on these five test set, our method achieves $3.3\%$, $6.3\%$, $1.3\%$, $1.9\%$ and $1.1\%$ improvement. The power of GDIFD for the DGOD task is shown.

(3) Compared to the state-of-the-art DGOD method DIDN, our DGOD framework with the same backbone achieves $3.4\%$, $0.8\%$, $2.6\%$ and $3.0\%$ in Foggy Cityscapes, Cityscapes, KITTI and BDD100k (C\&F\&K\&S to B), respectively. Our framework is inferior to DIDN by $0.7\%$ in the challenging setting of C\&F to B. It should be noted that our framework is much more lightweight compared to DIDN with multiple backbones in training, which is a notable advantage brought by our simple but effective DRL method. Results of DIDN with ResNet-101 are not provided in~\cite{didn}. We additionally provide the quantitative results of our framework with ResNet-101, which achieves the state-of-the-art performance in all the DGOD settings.

\subsubsection{Superiority of GDIFD to Other Lightweight DRL Methods for DGOD.}
Results of baselines equipped with different lightweight DRL methods are shown in Table~\ref{table:drl} to further prove the effectiveness of GDIFD. All the models with a DRL module are optimized via the same adversarial feature alignment losses, domain classification losses and detection losses. The orthogonal loss proposed in~\cite{vdd} is imposed on the other two DRL modules, which has a similar function to our gate loss to promote the orthogonality between DIR and DSR. 

The vanilla lightweight DRL methods proposed for UDAOD negatively affect the performance of the baseline model in DGOD settings, which is probably due to the fact that the absence of target training data make clean feature disentanglement mapping on the target domain very difficult to fit for the simple stacked convolutional layers in these methods. 
The model with VDD is even inferior to the model with the DRL method used in~\cite{liu2022decompose} according to our experiments, possibly because it becomes much more difficult to train a single encoder $E_{di}$ of weak representational capacity to directly extract high-quality DSR and DIR simultaneously in DGOD settings. 
Instead of directly extracting DIR from the input features, GDIFD only aims to output simple channel-wise gate signals, which also explicitly makes the backbone project DIR and DSR into different channels and enables much cleaner disentanglement of DIR from DSR on channel dimension. As shown in Table~\ref{table:drl}, only our lightweight DRL method outperforms the baseline model in these DGOD settings.

\setlength{\tabcolsep}{4pt}
\begin{table}
	\begin{center}
		\caption{Results (\%) of the baseline model with different lightweight DRL methods
		}
		\label{table:drl}
		\begin{tabular}{l|c|ccc}
			\hline \noalign{\smallskip} 
			Lightweight DGOD Method 			&Use DRL& B\&C to F  & B\&F to C & C\&F to B \\
			\noalign{\smallskip} \hline \noalign{\smallskip}
			Baseline-101			&	     				& 35.0  & 44.3 & 21.6 \\
			+ DRL in~\cite{liu2022decompose} (Fig.~\ref{fig:exist_disens}(b))&\checkmark & 34.6  & 43.5 & 19.2 \\
			+ VDD (Fig.~\ref{fig:exist_disens}(c))&\checkmark & 30.8 & 32.8 & 16.1 \\
			+ GDIFD (Ours, Fig.~\ref{fig:exist_disens}(d))&\checkmark & \textbf{38.3} & \textbf{50.6} &\textbf{22.9}\\
			\noalign{\smallskip}\hline 
		\end{tabular}
	\end{center}
\end{table}
\setlength{\tabcolsep}{1.4pt}

\subsection{Ablation Study}

\subsubsection{Effectiveness of the Proposed DGOD Method.}
We provide the results of different DGOD strategies in Table~\ref{table:alation} to validate the effectiveness of the proposed DRL method.
It is shown that only using the DSR learning part of GDIFD can enhance the model generalizability, since the domain-specific information can be largely separated out by DSR learning and the feature for object detection mainly encodes the task-relevant information. The additional adversarial learning can remove the domain-specific information in DIR and promote the independence between DIR and DSR as analyzed in Section~\ref{gdifd}, which further improves the DGOD performance of our framework. 
Additionally, our DRL method outperforms the non-disentangled method with only DIR learning, which can be mainly attributed to the fact that the alignment of DIR performs much better without interference from DSR due to the explicit domain-invariant feature disentanglement.
\setlength{\tabcolsep}{4pt}
\begin{table}
	\begin{center}
		\caption{Results (\%) of different DGOD schedules in the settings using B, C and F. DSRL: domain-specific representation learning in GDIFD with $\mathcal{L}_{D-cls}$ and $\mathcal{L}_{gate}$; DIRL: conventional adversarial domain-invariant representation learning with $\mathcal{L}_{D-adv}$}
		\label{table:alation}
		\begin{tabular}{l|cccc|cccc|cccc}
			\hline \noalign{\smallskip} 
			Setting   &\multicolumn{4}{c|}{B\&C to F} &\multicolumn{4}{c|}{B\&F to C} &\multicolumn{4}{c}{C\&F to B}\\
			\noalign{\smallskip} 
			\hline \noalign{\smallskip}
			& Base & & & Ours & Base & & & Ours & Base & & & Ours \\
			\noalign{\smallskip}
			\hline \noalign{\smallskip}
			DSRL& &\checkmark& &\checkmark & &\checkmark& &\checkmark & &\checkmark& &\checkmark   \\
			DIRL& & &\checkmark&\checkmark & & &\checkmark&\checkmark & & &\checkmark&\checkmark \\
			\noalign{\smallskip}
			\hline \noalign{\smallskip}
			mAP & 35.0 & 36.6 & 36.0  & \textbf{38.3} & 44.3 & 48.9 & 47.9 & \textbf{50.6}  & 21.6 & 21.7 & 22.1 & \textbf{22.9} \\	
			\noalign{\smallskip}	
			\hline
		\end{tabular}
	\end{center}
\end{table}
\setlength{\tabcolsep}{1.4pt}

\subsubsection{Effectiveness of the Configurations for CGM.}
\begin{figure}
	\centering
	\includegraphics[height=2.1cm]{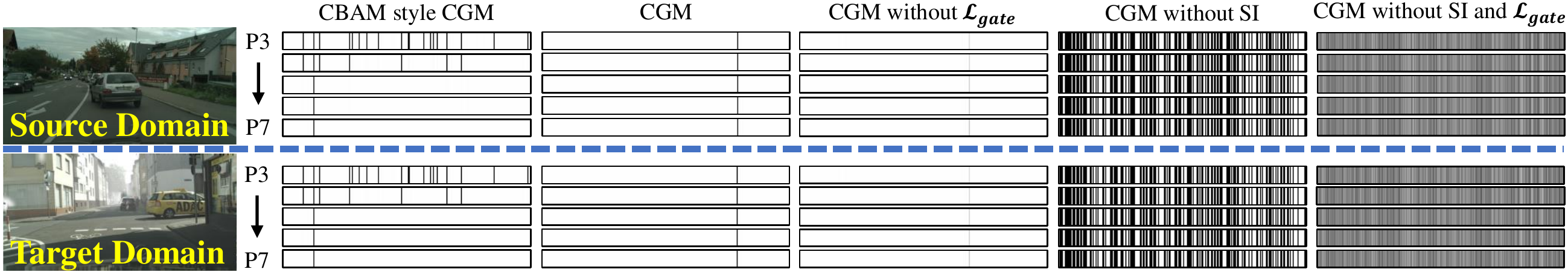}
	\caption{ $S_{di}$ for five scale levels from different channel gate modules (B\&C to F). $\mathcal{L}_{gate}$ and SI indicate the gate loss and the special initialization, respectively. Black for low domain-specific gate signals, and white for high domain-invariant gate signals}
	\label{fig:catt}
\end{figure}

We compare CGM to other modules of different configurations to demonstrate its effectiveness via the results in Table~\ref{table:signal}. The corresponding channel signals are visualized in Fig.~\ref{fig:catt}. As analyzed in Section~\ref{cgm}, the efficient channel attention module in CBAM~\cite{cbam} performs so rough global pooling before the gate signal estimation that it can not guarantee comparable results to CGM. Additionally, the visualization shows that the shared simple channel attention module of CBAM is not able to output consistent signals across the scale levels, which indicates that the backbone has to fit a more difficult mapping to project DSR of different levels to different channels and affects the DGOD performance negatively. 

\setlength{\tabcolsep}{4pt}
\begin{table}
	\begin{center}
		\caption{Results (\%) of our framework equipped with CGM of different configurations. $\mathcal{L}_{gate}$ and SI indicate using the gate loss and using special initialization, respectively}
		\label{table:signal}
		\begin{tabular}{l|cc|ccc}
			\hline \noalign{\smallskip} 
			Channel Gate Modules   & $\mathcal{L}_{gate}$  &\ SI\ \   & B\&C to F   & B\&F to C  & C\&F to B \\
			\noalign{\smallskip}
			\hline \noalign{\smallskip}
			CBAM style CGM  & \checkmark  &\checkmark   & 36.8  & 49.3 & 21.3\\
			CGM (Ours)  &\checkmark &\checkmark    & \textbf{38.3}  & \textbf{50.6} & \textbf{22.9} \\
			CGM  & \checkmark &  &  36.8  &  49.8 & 22.3 \\
			CGM  &  & \checkmark &  37.3 & 47.9 & 22.1 \\
			CGM  &  &  & 36.0  & 49.4 & 22.1 \\
			\noalign{\smallskip}
			\hline 
		\end{tabular}
	\end{center}
\end{table}
\setlength{\tabcolsep}{1.4pt}

The proposed CGM that uses both special initialization and the gate loss shows the best performance. 
On one hand, as shown in Fig.~\ref{fig:catt}, due to the gate signals initialized close to $1$ (channels are exclusive to DIR initially), the model will tend to use as few channels as possible for the domain classification task based on DSR. Specifically, our model learns to squeeze the task-irrelevant DSR to only one channel for the binary source domain classification in the setting of B\&C to F. Without the special initialization, the DGOD performance goes down, since too many channels are dedicated to encoding the task-irrelevant DSR and the extraction of the task-relevant DIR may be negatively affected. 
On the other hand, without the gate loss, the gate signal values are not constrained to either $0$ or $1$ and correspond to gray instead of black or white in the visualization. The gray gate signal values indicate that the corresponding channels are not exclusive to either DSR or DIR, which means that these two types of information that we want to disentangle on channel dimension are actually entangled in these channels. In a word, cleaner channel-wise disentanglement is not feasible without the gate loss, which leads to performance degradation.

\subsection{Extension to Domain Adaptation}
We also adapt our DGOD framework for the UDAOD task to further validate the effectiveness of our method. As shown in Table~\ref{table:udaod}, previous state-of-the-art UDAOD methods use different kinds of backbones, base detectors, base adversarial alignments and their own novel UDAOD approaches, which makes a fair comparison almost impossible. However, the effectiveness of our simple method can be validated by comparing the performance of our approach to the one of CFA, which has the same FCOS baseline as us. By using the proposed gated DRL method GDIFD instead of the additional center-aware feature alignment fine-tuning of CFA, our framework obtains $1.2\%$ improvement and shows good UDAOD performance. 
\setlength{\tabcolsep}{4pt}
\begin{table}
	\begin{center}
		\caption{Results (\%) of different methods in the UDAOD setting of C to F. Details of these methods are also described. Backbone VGG-16, ResNet-50, ResNet-101 are denoted by $v$, $r$ and $\hat{r}$, respectively. One-stage and two-stage object detection are denoted by $o$ and $\hat{o}$, respectively. Feature alignment on image-level and the one on both image-level and instance-level are denoted by $a$ and $\hat{a}$, respectively. Feature disentanglement on image-level and the one on both image-level and instance-level are denoted by $d$ and $\hat{d}$, respectively. 
			$\hat{g}$ denotes the graph-induced prototype alignment. 
			$\hat{f}$ denotes the additional model fine-tuning method. $\hat{c}$ denotes the categorical regularization method of VDD's backbone model. $\hat{u}$ denotes the uncertainty-aware method imposed on the adversarial alignment of image-level and instance-level. $\hat{m}$ denotes the metric learning methods on image-level and instance-level
		}
		\label{table:udaod}
		\begin{tabular}{llccccccccc}
			\hline \noalign{\smallskip} 
			Method  &Details   &person& car & train & rider & truck & motor & bike & bus & mAP \\
			\hline \noalign{\smallskip}
			GPA\cite{gpa} & $r\hat{o}\hat{a}\hat{g}$ & 32.9 & 54.1 & 41.1 & \textbf{46.7} & 24.7 & 32.4 & 38.7 & 45.7 & 39.5 \\
			CFA\cite{cfa}&    $\hat{r}oa\hat{f}$  		& 41.5 & 57.1 & 39.7 & 43.6 & 29.4 & 29.0 & 36.1 & 44.9 & 40.2 \\  
			VDD\cite{vdd}&  	$v\hat{o}\hat{a}d\hat{c}$   & 33.4 & 51.7 & 34.7 & 44.0 & \textbf{33.9} & \textbf{34.2} & 36.8 & 52.0 & 40.0\\  
			DIDN\cite{didn}& 	$r\hat{o}\hat{a}\hat{d}$	& 38.3 & 51.8 & 34.7 & 44.4 & 28.7 & 32.4 & \textbf{40.4} & \textbf{53.3} & 40.5\\
			UaDAN\cite{uadan} & $r\hat{o}\hat{a}\hat{u}$ & 36.5 & 53.6 & 42.7 & 46.1 & 28.9 & 32.3 & 38.9 & 49.4 & 41.1 \\
			DDF\cite{liu2022decompose}& 	$r\hat{o}a\hat{d}\hat{m}$		& 37.6 & 56.1 & \textbf{47.0} & 45.5 & 30.7 & 31.1 & 39.8 & 50.4 & \textbf{42.3}\\
			
			Ours&	$\hat{r}oad$		& \textbf{43.4} & \textbf{58.9} & 39.6 & 45.4 & 28.8 & 28.7 & 39.0 & 47.1 & 41.4 \\  
			\hline 
		\end{tabular}
	\end{center}
\end{table}
\setlength{\tabcolsep}{1.4pt}

\par\vfill\par
\section{Conclusions}
In this paper, we propose Gated Domain-Invariant Feature Disentanglement (GDIFD) for the DGOD task. To enhance the object detector generalizability on the unseen target domain, we disentangle domain-invariant feature more cleanly on channel dimension, which is used for the object detection task. Extensive experiments have been conducted to validate the effectiveness of GDIFD and the state-of-the-art performance of our framework. 
In the future, we will explore the power of GDIFD on other DG and UDA tasks.

\newpage
\bibliographystyle{IEEEtran}
\bibliography{IEEEabrv,../eccv2022/egbib}

\end{document}